\documentclass[runningheads]{llncs}

\usepackage{eccv}

\usepackage{eccvabbrv}

\usepackage{graphicx}
\usepackage{booktabs}

\usepackage[accsupp]{axessibility}  

\usepackage{hyperref}

\usepackage{orcidlink}

\usepackage{amsmath,amsfonts,bm}

\def\eqref#1{equation~\ref{#1}}

\def\1{\bm{1}}

\DeclareMathAlphabet{\mathsfit}{\encodingdefault}{\sfdefault}{m}{sl}
\SetMathAlphabet{\mathsfit}{bold}{\encodingdefault}{\sfdefault}{bx}{n}

\definecolor{mycyan}{RGB}{212, 239, 251}
\usepackage{colortbl}
\usepackage{xcolor}
\definecolor{mygray}{gray}{.9}
\definecolor{goldenrod}{RGB}{245,245,220}
\newlength\savewidth
\newcolumntype{a}{>{\columncolor{mygray}}c}
\usepackage{fontawesome5}
\usepackage{booktabs}
\usepackage{makecell}
\usepackage{fontawesome5}
\usepackage{lipsum}
\usepackage{comment}
\usepackage{multirow}
\usepackage{wrapfig}
\usepackage{algorithm}
\usepackage{algorithmic}
\usepackage{xfrac}  

\usepackage{graphicx}
\usepackage{color}

\definecolor{darkgreen}{rgb}{0,0.7,0}

\definecolor{mygraytext}{gray}{.5}

\def\ie{\emph{i.e.}}

\begin{document}

\title{Training-free Composite Scene Generation for Layout-to-Image Synthesis} 

\titlerunning{CSG for Layout-to-Image Synthesis}

\author{Jiaqi Liu\inst{1} \and
Tao Huang\inst{1} \and
Chang Xu\inst{1}}

\authorrunning{J.~Liu et al.}

\institute{School of Computer Science, Faculty of Engineering, The University of Sydney\\
\email{\{jliu6979,thua7590\}@uni.sydney.edu.au, c.xu@sydney.edu.au}
}

\maketitle

\begin{abstract}
    Recent breakthroughs in text-to-image diffusion models have significantly advanced the generation of high-fidelity, photo-realistic images from textual descriptions. Yet, these models often struggle with interpreting spatial arrangements from text, hindering their ability to produce images with precise spatial configurations. To bridge this gap, layout-to-image generation has emerged as a promising direction. However, training-based approaches are limited by the need for extensively annotated datasets, leading to high data acquisition costs and a constrained conceptual scope. Conversely, training-free methods face challenges in accurately locating and generating semantically similar objects within complex compositions. This paper introduces a novel training-free approach designed to overcome adversarial semantic intersections during the diffusion conditioning phase. By refining intra-token loss with selective sampling and enhancing the diffusion process with attention redistribution, we propose two innovative constraints: 1) an inter-token constraint that resolves token conflicts to ensure accurate concept synthesis; and 2) a self-attention constraint that improves pixel-to-pixel relationships. Our evaluations confirm the effectiveness of leveraging layout information for guiding the diffusion process, generating content-rich images with enhanced fidelity and complexity. Code is available at \url{https://github.com/Papple-F/csg.git}.

\end{abstract}

\begin{figure*}[t]
    \centering
    \includegraphics[width=\linewidth]{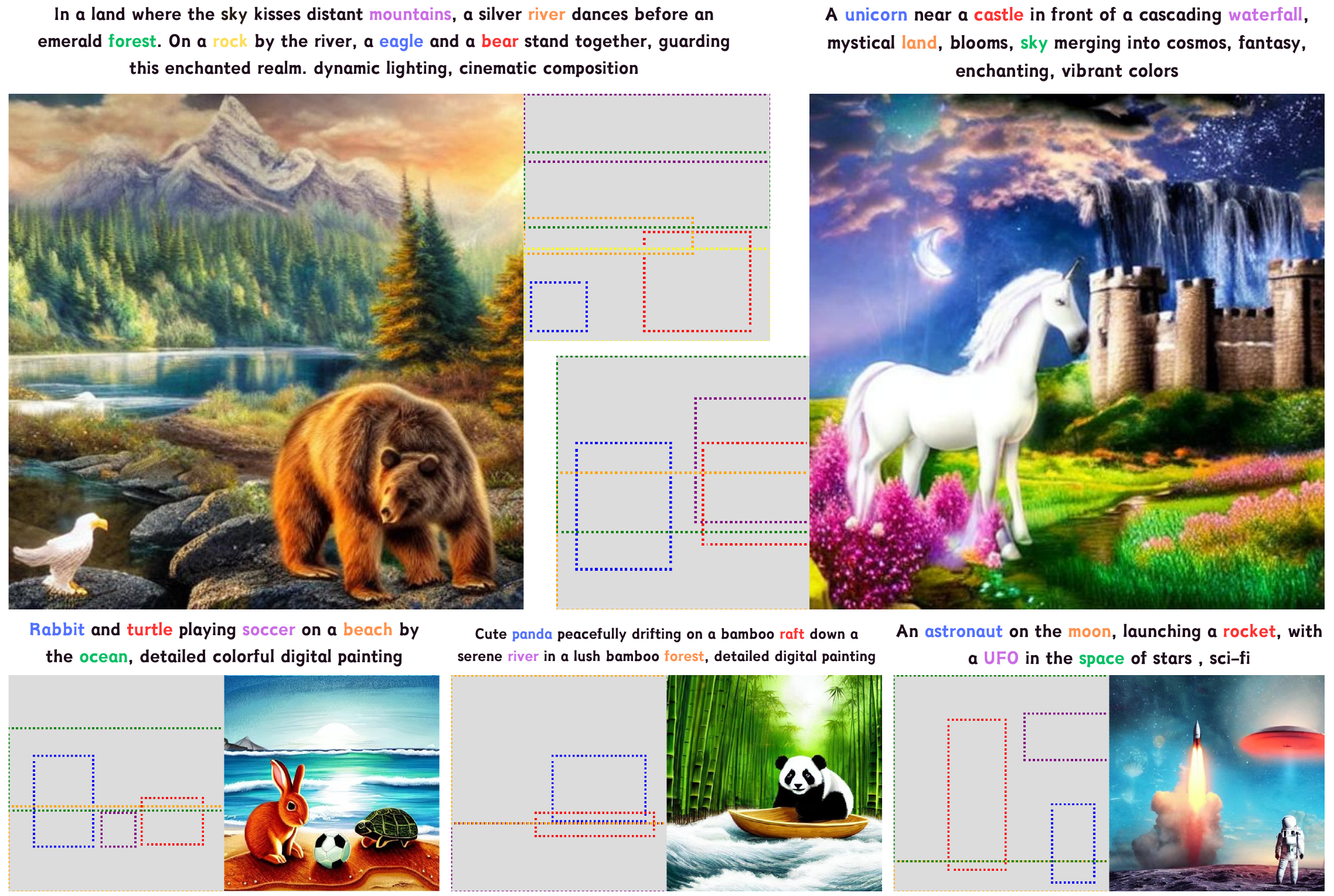}
    \caption{Composite scene generation facilitates the blending of various foreground and background elements into an image based on layout details.}
    \label{fig:art demo}
\end{figure*}

\section{Introduction}
\label{sec:intro}

Substantial advancements have been achieved in large-scale text-to-image generative models \cite{ramesh2021zero,Rombach_2022_CVPR,saharia2022photorealistic,shi2020improving,ramesh2022hierarchical,rombach2022high}, which are now capable of converting complex text descriptions into visually compelling images with impressive accuracy and creativity. Despite these advancements, current models still struggle to comprehend layout descriptions contained within texts and to precisely outline and render detailed images. This is largely due to limitations in model capacity and the quality of training datasets. Consequently, in tasks such as art creation \cite{dehouche2023s} and industrial design \cite{liu2023application}, where elements need to be precisely positioned, the models' intrinsic inability to comprehend user-defined spatial information limits their applicability.

Recent advancements in layout-to-image approaches have shown significant progress in achieving more controllable generation, employing both training-based \cite{li2021image,gafni2022make,zheng2023layoutdiffusion,chai2023layoutdm,balaji2022ediffi,zhang2023adding,li2023gligen} and training-free \cite{xie2023boxdiff,kim2023dense,chen2024training,bar2023multidiffusion} methodologies to transform detailed layouts into vivid, realistic images. While training-based methods have demonstrated promise, their effectiveness heavily relies on the availability of comprehensive and varied layout datasets. Assembling, annotating, and preprocessing these datasets for effective concept learning is challenging due to the high costs and resource-intensive nature of data acquisition, posing a substantial barrier for many research initiatives. In contrast, training-free methods utilize the intrinsic generative capabilities of models to dynamically guide the diffusion process, offering a significant efficiency advantage. As text-to-image models become more popular, there is an increasing demand for content generation that involves complex compositions of multiple objects and backgrounds. Existing layout-to-image approaches, which primarily focus on content positioning, often struggle with conflicts arising from intersecting similar concepts, especially when generating scenes with multiple elements. Moreover, the critical role of self-attention mechanisms in refining generated content with accurate textural details during the later stages of diffusion, as highlighted in \cite{balaji2022ediffi}, is often overlooked. This oversight highlights a gap in current methodologies, which may fail to fully leverage the potential of diffusion models for producing detailed and contextually coherent images.

This study aims to generate composite scenes featuring multiple objects and backgrounds, utilizing bounding boxes for layout information. We propose enhancing the training-free backward guidance concept \cite{chefer2023attend,chen2024training} with a novel selective sampling strategy. This strategy introduces a dropout mechanism that prioritizes attentions closely aligned with the current generation concept during the calculation of intra-token constraints. Such an approach not only improves content positioning accuracy but also ensures broader coverage of the targeted area, addressing key challenges in layout-to-image generation.

Our approach extends beyond individual token cross-attentions by implementing an inter-token constraint, evaluating attentions across tokens within a targeted region to ensure the prioritization of the correct concept. This technique aims to counteract semantic intersection—where overlapping concepts produce irregular textures and shape inaccuracies—thereby enhancing generation precision and reliability. Recognizing adversarial intersections' impact on pixel relationships, we employ a self-attention constraint for collective self-attention adjustment across the target region, critical in the later diffusion stages for maintaining coherent pixel interactions. Additionally, an attention redistribution method during forward diffusion corrects misaligned attentions, addressing refinement limitations and reducing semantic intersection effects, thereby improving overall generation accuracy.

Our comprehensive experimental evaluations demonstrate that our method significantly outperforms existing training-free layout-to-image generation approaches in terms of content localization accuracy and semantic fidelity. These advancements are vividly illustrated in the examples presented in Figure \ref{fig:art demo}, showcasing our method's ability to maintain higher semantic correctness while accurately positioning content within the generated images.

\section{Related Work}
\label{sec:related}
\subsection{Text-to-image generative models}

Recent advancements in text-to-image generative models, exemplified by Stable Diffusion \cite{Rombach_2022_CVPR}, DALLE-3 \cite{shi2020improving}, and Imagen \cite{saharia2022photorealistic}, represent a significant leap forward beyond previous dominant techniques like generative adversarial networks \cite{goodfellow2014generative,karras2019style,brock2018large,guo2020positive}. These new models excel not only in image generation but also in enhancing the performance of tasks such as classification \cite{huang2024active,azizi2023synthetic}, action segmentation \cite{liu2023diffusion,ding2023temporal}, and more. These models enable the generation of highly detailed visual content directly from textual descriptions, distinguishing themselves by producing contextually relevant and aesthetically pleasing images. Their ability to translate complex textual prompts into visual artworks demonstrates a remarkable proficiency, unlocking new possibilities across various applications, from digital art creation \cite{dehouche2023s} to diverse content generation.

However, these models heavily rely on large-scale datasets (stable diffusion is trained using billions of images from LAION-5B \cite{schuhmann2022laion}), which poses limitations in extending their capabilities to cover new tasks with learned concepts without incorporating additional task-specific datasets or extensive training. As previous studies have showcased the feasibility of integrating various novel tasks in a plug-and-play manner, including image editing \cite{hertz2022prompt,parmar2023zero,tumanyan2023plug,cao2023masactrl}, generation enhancement \cite{chefer2023attend,li2023divide}, layout-to-image \cite{xie2023boxdiff,kim2023dense,chen2024training,bar2023multidiffusion}, and more. Encouraged by these findings, we are motivated to delve deeper into whether the boundaries of layout-to-image tasks can be expanded further, enabling the model to handle scenarios involving multiple objects and backgrounds.

\subsection{Layout-to-image generation}

The reliance on purely linguistic methods limits a model's ability to decode specific layout details accurately. Various studies \cite{li2021image,gafni2022make,zheng2023layoutdiffusion,chai2023layoutdm,balaji2022ediffi,zhang2023adding,li2023gligen} have shown that models can be further trained with layout information to facilitate layout-to-image generation tasks. Nonetheless, these approaches also highlight the challenge of requiring datasets that pair image, text, and layout information—resources that are scarce and costly to compile. Some innovative strategies have aimed to integrate layout details without additional model training. For instance, MultiDiffusion \cite{bar2023multidiffusion} involves denoising and combining regions with corresponding text descriptions, while DenseDiff \cite{kim2023dense} directly modifies the attention probabilities to enhance focus on the targeted regions. BoxDiff \cite{xie2023boxdiff} and Layout-control \cite{chen2024training} both employs generative semantic nursing \cite{chefer2023attend}, which optimizing latents based on cross-attentions to achieve desired layouts.

Our approach adopts the backward guidance framework utilized by BoxDiff \cite{xie2023boxdiff} and Layout-control \cite{chen2024training}. However, while these existing methods are efficient for straightforward situations, their method narrowly focus on cross-attentions for individual tokens, overlooking the potential for semantic overlaps as the layout becomes more complex with the addition of multiple objects. This oversight can lead to the undesirable blending of features and disrupt the integrity of pixel relationships, which may prevent the accurate generation of targeted objects. Unlike these methods, our strategy takes a comprehensive perspective on the optimization process, integrating considerations of intra-token, inter-token cross attentions, and self-attentions. This broader approach demonstrates that a training-free method can not only adhere more closely to the intended layout but also enhance the quality of the generated images.

\section{Preliminaries}
\label{sec:preliminaries}
\subsection{Stable diffusion}

Diffusion models \cite{sohl2015deep,ho2020denoising,croitoru2023diffusion,dhariwal2021diffusion}, operates by gradually transforming a random noise distribution into a coherent image, guided by the semantics of the input text. This process involves a series of forward and backward steps, where the model initially adds noise $\epsilon$ to an image $\bm{\mathcal{I}}$ and then learns to recover the original image from noise, conditioned on textual descriptions $\bm{c}$. Latent diffusion model \cite{Rombach_2022_CVPR} operates in latent space where for a given image $\bm{\mathcal{I}}$, it is first encoded as latent $\bm{z}$ by an encoder $\mathcal{E}$, and then reconstructed by a decoder $\mathcal{D}$ as $\hat{\bm{\mathcal{I}}} = \mathcal{D}(\bm{z}) = \mathcal{D}(\mathcal{E}(\bm{\mathcal{I}}))$ after the denoising process.

During the training stage of diffusion process, at given timestep $t$, with latent $\bm{z}_t$ and condition $\bm{c}$, the denoiser $\epsilon_\theta$ learns to correctly predict the added noise $\epsilon$ through mean square error:
\begin{equation}
    \mathcal{L}_{diff} = \mathbb{E}_{z\sim\mathcal{E}(\bm{\mathcal{I}}),\epsilon\sim N(0,1),\bm{c},t} [\|\bm{\epsilon} - \bm{\epsilon}_\theta(z_t,c,t)\|^2_2].
\end{equation}

During the inference stage, the denoising process gradually removes noise from a randomly sampled $\bm{z}_T$ with the noises predicted by diffusion model with condition $\bm{c}$, and guide the diffusion process through classifier \cite{dhariwal2021diffusion} or classifier-free guidance \cite{ho2022classifier}.

\subsection{Attention mechanism}

The attention mechanism forms the cornerstone of transformer models \cite{vaswani2017attention}, playing a pivotal role in natural language processing (NLP) tasks. This mechanism was subsequently integrated into the vision transformer architecture \cite{dosovitskiy2020image}, thereby extending its applicability to tasks within the realm of computer vision. Within the framework of stable diffusion models, the intermediate features are processed through both self-attention and cross-attention layers. This processing facilitates the generation of an attention map at timestep $t$ as follows:
\begin{equation}
    A_t = softmax(\frac{QK^T}{\sqrt{d}}).
\end{equation}

In this process, Q and K represent queries and keys, respectively, with d denoting the dimensionality of these query and key features. For cross-attention mechanisms, K is derived from the projection of text embeddings. These embeddings result from encoding the text conditioning $c$ into a latent space, achieved through the use of a CLIP text encoder \cite{radford2021learning}. Conversely, Q corresponds to the projection of intermediate features sourced from a U-Net architecture. Within the self-attention framework, both Q and K are obtained from the projections of these intermediate features.

\section{Methods}

In this section, we explore the nuances of generating composite scenes. We begin in Section \ref{sec:cc} by detailing our novel approach to attention selection, including the definition of mask representations and the application of our selective sampling strategy. This section also explains the development and role of intra- and inter-token constraints in influencing the diffusion process. In Section \ref{sec:sc}, we shift focus to our innovative handling of self-attentions, illustrating their pivotal role in refining the diffusion process for enhanced image generation. Finally, Section \ref{sec:opt} introduces our attention redistribution technique, employed during forward diffusion to further optimize generation outcomes through strategic constraint refinement and application.

\begin{figure*}[t]
    \centering
    \includegraphics[width=\linewidth]{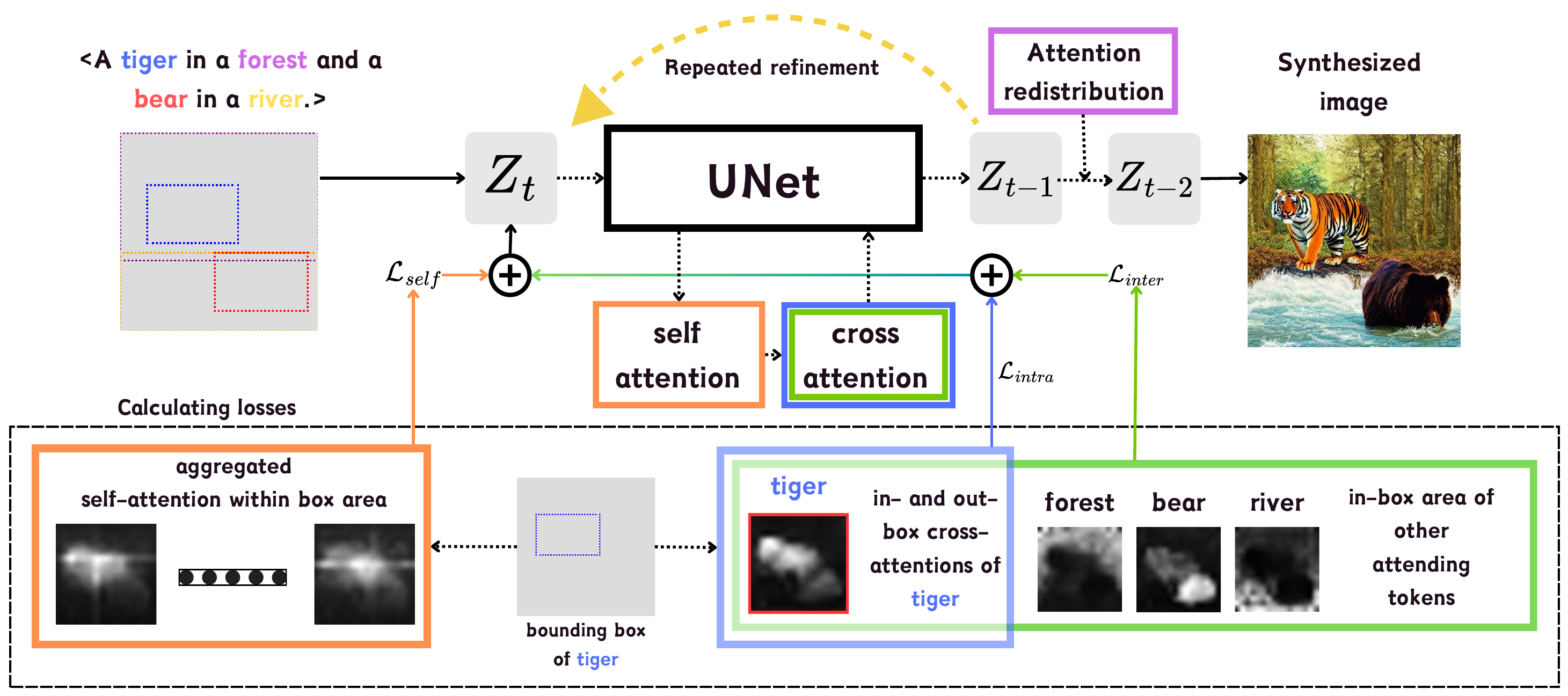}
    \caption{In the workflow of Composite Scene Generation (CSG), at each refinement stage, we capture both self- and cross-attentions within a UNet structure. For self-attention, we aggregate self-attentions within each mask area $\bm{m}_i$ and calculate $\mathcal{L}_{self}$, which determines if pixel-level interaction is mostly constrained within the target area. For cross-attention, we first obtain $\mathcal{L}_{intra}$, a proportional measure of in-box and out-box cross-attentions for each attending token. Next, we assess the cross-attentions for all attending tokens within the same box area to obtain $\mathcal{L}_{inter}$, determining if the cross-attention of the current token is dominant within its own region. After a finite number of refinement steps, the latent is updated through the gradient of all three loss components. To further enhance the refinement process, we implement attention redistribution between each refinement stage.}
    \label{fig:workflow}
\end{figure*}

\subsection{Cross-attention constraints}
\label{sec:cc}

To formulate constraints on cross-attentions, we begin by detailing the acquisition of attentions and layout information for constraint application. Leveraging latent noise manipulation is pivotal in aligning image-text attentions within designated spatial regions. Drawing on recent research \cite{chefer2023attend,xie2023boxdiff} which suggests that attentions at smaller scales capture semantic information more effectively during early diffusion phases, we employ averaged cross-attentions from user-defined attending objects at the $16 \times 16$ up-scaling block as our manipulation references. Formally, for $N$ attending tokens, we define $\mathcal{A} = \{\bm{A}_i\}_{i=1}^{N}, \bm{A}_i \in \mathbb{R}^{L}$ as the set of cross-attentions relative to the tokens, with $L$ indicating the number of image tokens in cross-attentions.

For representing layout information, we employ bounding boxes $\bm{\mathcal{B}} = \{\bm{b}_i\}_{i=1}^N$, which delineate the top-left and bottom-right corner coordinates for each object. Correspondingly, each bounding box $\bm{b}_i$ is transformed into a binary mask $\bm{m}_i$, with pixels inside the box region marked as 1 and those outside as 0.

The mask is leveraged in our constraints to discriminate foreground and background pixels of the generation of each object. To mitigate the potential disruption of latent noise's natural Gaussian distribution by gradient-based backpropagation, while ensuring extensive attention coverage within the target region, we employ a \texttt{selective sampling} strategy $\mathcal{F}(\bm{m_i})$ for each mask $\bm{m_i}$. Specifically, we select the top $K$ elements within the mask region for focused attention. To avoid concentrating excessively on a limited number of elements, we randomly retain $M$ elements from those selected. 

$\textbf{Intra-token constraint:}$ For each attending token, we use its corresponding mask $\bm{m_i}$ and
cross-attention probabilities $\bm{A}_i$ to compute our intra-token attention regularization loss. To make sure that the object is only generated in its own target region (\ie, inside the bounding box), we encourage the cross-attention values associated with the target region to exceed those of the non-target region. Following \cite{chen2024training}, we define the intra-token constraint as:
\begin{align}
    \begin{split}
        \mathcal{L}_{intra} &= \sum_{i=1}^N{\mathcal{L}^{intra}_i}\\
        \mathrm{with}\quad \mathcal{L}^{intra}_i &= (1 - \frac{\sum_{j=1}^L{(\tilde{\bm{m}}_i\cdot\bm{A}_i)_j}}{\sum_{j=1}^L{(\tilde{\bm{m}}_i\cdot\bm{A}_i)_j} + \sum_{j=1}^L{(\hat{\bm{m}}_i\cdot\bm{A}_i)_j}})^2,
    \end{split}
\end{align}
where $\tilde{\bm{m}}_i = \mathcal{F}(\bm{m}_i)$ and $\hat{\bm{m}}_i = \mathcal{F}(1-\bm{m}_i)$.

The constraint is designed to increase the proportion of attention values within the target region relative to those outside, enhancing focus on the intended area. By applying selective sampling to pivotal values, we address the issue of excessive attention concentration in confined areas. Furthermore, selective sampling from both inside and outside regions ensures a more equitable representation of high values, mitigating bias in the aggregation process and enhancing the fairness of their contribution to the loss function.

$\textbf{Inter-token constraint:}$ The semantic intersection due to the conceptual similarity (\textit{e.g.} a bear and a tiger shares considerable similarity in textual latent space), under the context of limited latent space, inevitably intertwine their cross-attentions and cause a degree of spatial intersection. Aiming to completely eliminate such adversarial effects may not be plausible due to the inherent dependence between image intermediate features are extracted from the same latent variable and difficulties to manipulate textual latent space without altering its original semantic meanings. Thus, we employ a conservative approach where we encourage attentions inside corresponding region not to be exclusive, but surpass over other tokens' attentions at the same location. To accomplish this, for a given token $i$, we first obtain the maximum attentions within target region for other tokens:

\begin{equation}
    (\tilde{\bm{m}}_i\cdot\bm{A}_j)^{max} = \max_{j=1,j\ne i}^N(\tilde{\bm{m}}_i\cdot\bm{A}_j).
\end{equation}

Following, we calculate the difference $d$ between such attentions and attentions of target region of given token with a margin $g$.
\begin{equation}
    d = \sum_{k=1}^{L}((\tilde{\bm{m}}_i \cdot \bm{A}_{i})_k - g) - \sum_{k=1}^L(\tilde{\bm{m}}_i \cdot \bm{A}_{j})_k^{max}.
\end{equation}

To achieve our goal of making the given token more prominent than others within the targeted region, we establish an inter-token constraint defined as follows:
\begin{equation}
\mathcal{L}^{inter}_i = 
\left\{
\begin{aligned}
  & 0, d > 0 \\
  & d^2, d < 0
\end{aligned}
\right.,\
\mathcal{L}_{inter} = \sum_{i=1}^N{\mathcal{L}^{inter}_i}.
\end{equation}

The constraint on inter-token relations, utilizing the same elements from a given $\bm{m}_i$, focuses on a side-by-side comparison of cross-attention values with other tokens at identical spatial locations. This optimization is particularly aimed at scenarios involving intersections, encompassing both the semantic similarities previously discussed and intersections of user-specified regions. The goal is to promote the dominance of a single token's content within a specific area to ensure the generated image is both clear and coherent.

\subsection{Self-attention constraints}
\label{sec:sc}

The impact of cross-attention is critical in the initial diffusion stages, guiding the structural formation of content via textual cues. However, as \cite{balaji2022ediffi} highlights, self-attention gains prominence in later diffusion stages, influencing object textures. An unintended effect of semantic intersection can misguide self-attention, causing inappropriate pixel correlations during early stages. To mitigate this, we propose to align self-attention within target regions in early diffusion stages, improving local coherence and maintaining necessary external interactions. 

Following the same setup as section \ref{sec:cc}, we obtain averaged self-attentions $\bm{S} \in \mathbb{R}_{>0}^{L\times L}$ at 16 $\times$ 16 up-scaling blocks. Given mask $\bm{m}_i$, we aggregated self-attentions which lies within the mask region:
\begin{equation}
    \bm{S}_i = \sum_{j=1}^{L} (\bm{m}_i \cdot \bm{S})_j,
\end{equation}
where $\bm{S}_i \in \mathbb{R}_{>0}^L$.

Similar to the intra-token constraint, the self-attention constraint is defined as:
\begin{align}
    \begin{split}
        \mathcal{L}_{self} &= \sum_{i=1}^N{\mathcal{L}^{self}_i}\\
        \text{with}\quad \mathcal{L}^{self}_i &= (1 - \frac{\sum_{j=1}^L{(\tilde{\bm{m}}_i\cdot\bm{S}_i)_j}}{\sum_{j=1}^L{(\tilde{\bm{m}}_i\cdot\bm{S}_i)_j} + \sum_{j=1}^L{(\hat{\bm{m}}_i\cdot\bm{S}_i)_j}})^2,
    \end{split}
\end{align}
where $\tilde{\bm{m}}_i = \mathcal{F}(\bm{m}_i)$ and $\hat{\bm{m}}_i = \mathcal{F}(1-\bm{m}_i)$.

The self-attention constraint is based on the assumption that pixels within an object have stronger connections to other pixels associated with the same object. Therefore, through selective sampling of averaged self-attention scores, we reinforce these connections by emphasizing self-attentions that are most relevant to the object. This method also leverages the principle that total self-attention probabilities equal to 1 along each channel, meaning while enhancing connections between pixels linked to the object, less relevant pixels are diverted outside the target area, thereby improving the object's coherence with its surroundings.

\subsection{In-generation diffusion guidance}
\label{sec:opt}

Given that the limited latent space can cause semantic-level overlaps, this results in unresolved spatial overlaps even with sufficient refinement steps. These errors accumulate during diffusion steps, potentially leading the generation results in incorrect directions. To address this, we introduce attention redistribution, a technique that reallocates cross-attentions with corresponding tokens during the diffusion process. For each token, its cross-attention is defined as the aggregated cross-attentions across all attending tokens within the bounding box area. After reallocation, we apply max normalization to all cross-attentions. The attention redistribution is defined as follows:

\begin{equation}
    \bm{A}_i = \bm{m}_i\cdot\sum_{j=1}^N(\bm{A}_j)
\end{equation}

The generative process is divided into refinement and guidance stages. During each guidance step, the latent variable is updated $T_R$ times (the number of refinement steps per timestep). This process lasts for $T_D$ steps (the total number of timesteps during the diffusion process with such recurrent updates).

For each refinement step, the overall constraint is defined as:
\begin{equation}
    \mathcal{L} = \mathcal{L}_{intra} + \mathcal{L}_{inter} + \mathcal{L}_{self},
\end{equation}
and the latent is updated with a linear decay factor $\eta_t$ as:
\begin{equation}
    \bm{z}' = \bm{z}_t - \eta_t \cdot \nabla_{\bm{z}_t} \mathcal{L}.
\end{equation}

Through this approach, at each step of the diffusion process, we refine the latent to achieve more focused attention on the targeted regions. This refinement process helps to resolve conflicts among tokens and aligns self-attentions, thereby improving both internal and external connections within the generated scenes.

\section{Experiments}
\subsection{Experimental setup}
\label{exp:imp}
\textbf{Evaluation: }In our quantitative analysis using the COCO 2014 dataset \cite{lin2014microsoft}, we adopt a methodology akin to \cite{bar2023multidiffusion}, filtering the dataset to include images with n distinct objects where each occupies at least 5\% of the image area. We further refine the selection by excluding objects described by more than two words, as well as ``person'', resulting in a final set of 61 object classes. For performance evaluation, we employ YOLOv7 \cite{wang2023yolov7} for object detection, utilizing metrics such as YOLO-score (AP, AP$_{50}$) \cite{li2021image} to assess our method's effectiveness of locating and correctly generating objects. Object detection is conducted across all 80 COCO classes to ensure thoroughness. Additionally, we generate randomly sample and generate approximately 1,000 samples for subsets of the COCO 2014 dataset, categorized by the presence of 2, 3 and 2-4 distinct objects, using prompts structured as ``\texttt{a \{object$_1$\} ... and a \{object$_n$\}}'' to systematically evaluate our approach.

We utilize CLIP-score \cite{radford2021learning} for quantitatively assessing image-text compatibility, thereby evaluating the semantic accuracy of synthesized images. The text descriptions for images are formulated by appending a prefix of "a photo of" to the generation prompts.

\textbf{Implementation: }The experimental findings were derived using the CompVis1.4 model for text-to-image synthesis. We configured the model to perform 50 denoising steps with a constant guidance scale of 7.5, and produced synthetic images at a resolution of 512 × 512. The hyperparameters of selective sample is set to select the highest 80\% of attention values and randomly keep 50\%. the margin for the inter-token constraint was established at 0.1, a setting that resulted in achieving the highest scores for both YOLO and CLIP metrics.

\begin{table}[t]
    \centering
        \small
        \setlength{\tabcolsep}{3mm}
        \renewcommand{\arraystretch}{1.2}
        \caption{Ablation study on proposed components.}
        \label{tab:ab_guidance}
        \begin{tabular}{ccccc|ccc}
            \toprule
            $\mathcal{L}_{intra}$ & $\mathcal{L}_{inter}$ & $\mathcal{L}_{self}$ & SS & AR & AP$_{50}$ & AP & CLIP score\\
            \midrule
            \checkmark &&& \checkmark & \checkmark & 32.1 & 11.4 & 0.3202\\
            \checkmark & \checkmark &&&& 35.5 & 12.5 & 0.3196\\
            \checkmark & \checkmark && \checkmark & \checkmark & 40.2 & 16.1 & 0.3164\\
            \checkmark & \checkmark && \checkmark && 38.1 & 14.5 & 0.3175\\
            \checkmark & \checkmark &&& \checkmark & 39.1 & 14.3 & \textbf{0.3206}\\
            \checkmark & \checkmark & \checkmark & \checkmark &  & 46.8 & 19.3 & 0.3200\\
            \checkmark & \checkmark & \checkmark & \checkmark & \checkmark & \textbf{50.3} & \textbf{20.4} & 0.3201\\
            \bottomrule
        \end{tabular}
\end{table}

\subsection{Case studies}

\textbf{Quantitative ablation studies: }The comprehensive evaluation of our proposed methods encompasses $\mathcal{L}_{intra}$, $\mathcal{L}_{inter}$, $\mathcal{L}_{self}$, selective sampling, and cross-attention redistribution, as summarized in the table \ref{tab:ab_guidance}. The results reveal that utilizing only $\mathcal{L}_{intra}$ with selective sampling and cross-attention redistribution yields subpar object generation and location, achieving a mere 32.1 AP$_{50}$ and 11.4 AP. Introducing $\mathcal{L}_{inter}$ to emphasize the cross-attention of the token over others at the same location leads to a significant enhancement, with improvements of 7.9 in AP$_{50}$ and 3.7 in AP. This highlights the presence of attention overlap, as previously mentioned, and underscores the effectiveness of our method in addressing this issue. The incorporation of $\mathcal{L}_{self}$ further boosts performance, resulting in 10.1 AP$_{50}$ and 4.3 AP. This suggests that the presence of self-attention distraction compromises the positioning accuracy of layout-to-image methods. The concept of aggregating and treating self-attentions as a whole emerges as an effective strategy to mitigate such issues. Selective sampling and cross-attention redistribution also contribute quantitatively to improved efficacy.

In terms of the CLIP score, the introduction of $\mathcal{L}_{inter}$ appears to compromise the coherence of image semantics, causing a slight reduction from 0.3202 to 0.3164. However, with the addition of $\mathcal{L}_{self}$, our strategy of enhancing interconnection with object pixels while maintaining sufficient redundancy with other pixels proves to enhance semantic coherence.

\begin{figure*}[t]
    \centering
    \includegraphics[width=\linewidth]{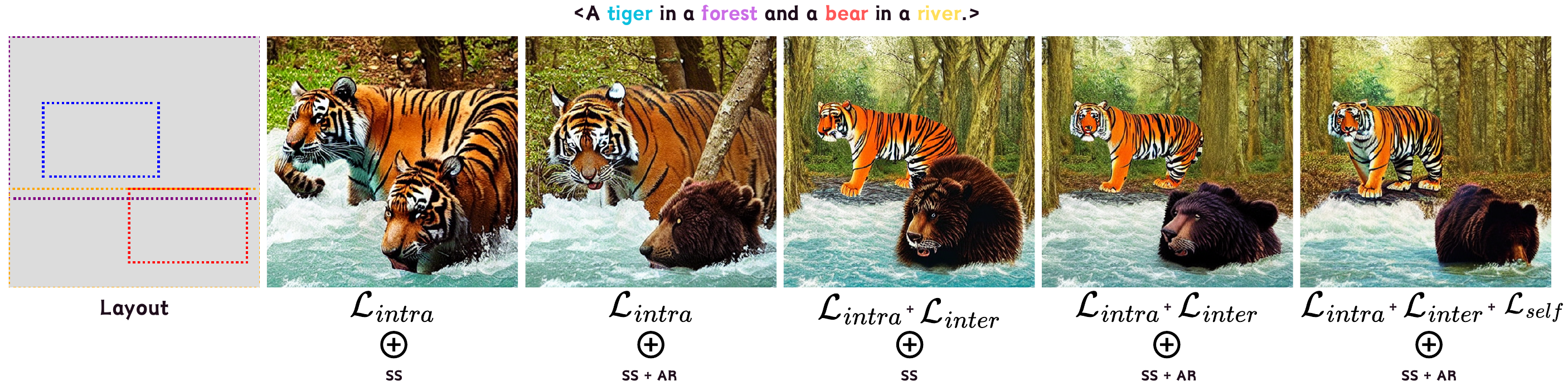}
    \caption{Visual ablation studies on various components of proposed method.}
    \label{fig:vis_ab}
\end{figure*}

\textbf{Qualitative ablation studies: }We conducted visual ablation studies to elucidate the progressive impact of our method on the generation process. In the latent space of textual representations, two objects, namely \texttt{a tiger and a bear}, inherently share similarities. According to figure \ref{fig:vis_ab}, when only controlling $\mathcal{L}_{intra}$, these two concepts inevitably overlap and merge, with the more dominant concept exerting greater influence over the generated content. Introducing $\mathcal{L}_{inter}$ and AR helps mitigate this overlap to some extent. While $\mathcal{L}_{inter}$ enhances control over layout precision by competing for the same location, it does violate the fidelity of the resulting image. On the other hand, AR provides a more natural transition from objects to backgrounds. Finally, with the addition of $\mathcal{L}_{self}$, we further fine-tune location precision and re-establish the relationship between objects and backgrounds, aiming to recover some of the fidelity lost due to the impact of $\mathcal{L}_{inter}$.

\begin{table}[t]
    \centering
    \small
    \setlength{\tabcolsep}{3mm}
    \renewcommand{\arraystretch}{1.2}
    \caption{Ablation study on margin with K = 80\%, M = 50\%.}
    \label{tab:ab_margin}
    \begin{tabular}{c|ccc}
        \toprule
        margin & AP$_{50}$ & AP & CLIP score\\
        \midrule
        0.05 & 48.8 & 19.3 & 0.3179\\
        0.1 & \textbf{50.3} & \textbf{20.4} & \textbf{0.3201}\\
        0.2 & 46.8 & 19.8 & 0.3167\\
        \bottomrule
    \end{tabular}
\end{table}

\textbf{Hyperparameter ablation studies: } Hyperparameters of our work, including $K$ and $M$ for selective sampling and the margin in the inter-token constraint, play a critical role in the effectiveness of the proposed components and collectively impact the overall performance of our method. The detailed results of hyperparameter optimization are presented in Tables \ref{tab:ab_margin}, \ref{tab:ab_K}, and \ref{tab:ab_M}. Specifically, for the inter-token constraint margin, as shown in Table \ref{tab:ab_margin}, a margin of 0.1 provides the best results. This finding is somewhat counter intuitive, as one might expect that a larger margin would more effectively differentiate between token attentions. However, as discussed in Section \ref{sec:cc}. A plausible explanation for this observation might lie in the fundamentally intertwined nature of cross-attentions that intersect, influenced by text embeddings confined to a constrained latent space, preventing concepts from being represented distinctly, and the latent features that are shared among all tokens.

\begin{table}[t]
    \centering
    \begin{minipage}[c]{0.49\linewidth}
        \centering
        \small
        \setlength{\tabcolsep}{2mm}
        \renewcommand{\arraystretch}{1.2}
        \caption{Ablation study on K with margin = 0.1, M = 50\%.}
        \label{tab:ab_K}
        \begin{tabular}{c|ccc}
            \toprule
            K & AP$_{50}$ & AP & CLIP score\\
            \midrule
            100\% & 46.6 & 18.5 & 0.3194\\
            80\% & \textbf{50.3} & \textbf{20.4} & 0.3201\\
            60\% & 43.6 & 16.0 & 0.3200\\
            40\% & 37.2 & 13.3 & \textbf{0.3207}\\
            \bottomrule
        \end{tabular}
    \end{minipage}\hfill
    \begin{minipage}[c]{0.49\linewidth}
        \centering
        \small
        \setlength{\tabcolsep}{2mm}
        \renewcommand{\arraystretch}{1.2}
        \caption{Ablation study on M with margin = 0.1, K = 80\%.}
        \label{tab:ab_M}
        \begin{tabular}{c|ccc}
            \toprule
            M & AP$_{50}$ & AP & CLIP score\\
            \midrule
            100\% & 48.2 & 17.7 & \textbf{0.3201}\\
            75\% & 48.9 & 19.1 & \textbf{0.3201}\\
            50\% & \textbf{50.3} & \textbf{20.4} & \textbf{0.3201}\\
            25\% & 45.7 & 16.5 & 0.3197\\
            \bottomrule
        \end{tabular}
    \end{minipage}
\end{table}

According to table \ref{tab:ab_K}, selecting a high-value portion with $K=80\%$ results in optimal performance for both AP$_{50}$ and AP metrics, though there is a slight decrease in CLIP score. This can be attributed to the fact that optimizing latents through gradient-based backpropagation can disrupt their naturally Gaussian distribution, potentially compromising image fidelity if not managed cautiously. With a smaller $K$, our method affects only a limited number of attention values, minimally directing the diffusion process. This approach preserves semantic integrity to some extent but sacrifices precision in object placement and generation. Conversely, using a higher $K$ value, while including random drops, approximates the selection of highly-related attention values to random sampling, which harms performance.

Achieving the best performance with a margin of 0.1 and $K$ at 80\%, we find optimal results at $M$ equal to 50\% (table \ref{tab:ab_M}). This outcome aligns with the patterns observed in our $K$ ablation studies, indicating a balanced exists in our selective sampling technique that prohibits incorporating either too many or too few attention values.

\subsection{Qualitative analysis}

\begin{figure*}[t]
    \centering
    \includegraphics[width=0.92\linewidth]{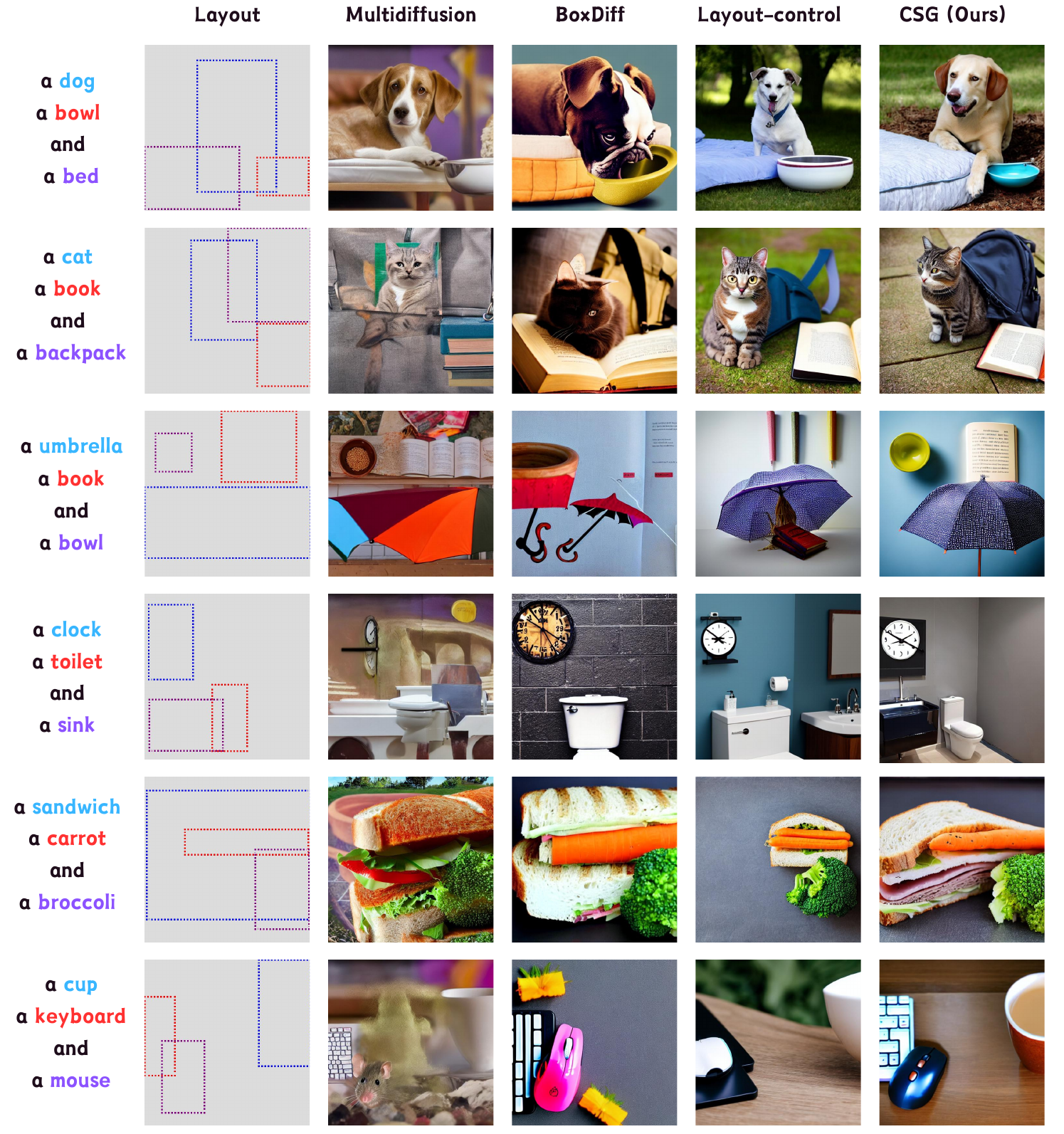}
    \caption{Visual comparison with concurrent training-free methods including MultiDiffusion \cite{bar2023multidiffusion}, BoxDiff \cite{xie2023boxdiff} and Layout-control \cite{chen2024training}. Layout information is sampled from COCO \cite{lin2014microsoft} with 3 distinct objects.}
    \label{fig:vis_cmp}
\end{figure*}

When conducting a visual comparison (figure \ref{fig:vis_cmp}) using images generated from datasets featuring three distinct objects, we observe that previous methods often struggle with producing a coherent image representation or accurately placing objects. For example, while MultiDiffusion excels at positioning objects in specified regions, it frequently compromises the overall image coherence and object integrity, leading to visually jarring images with poor transitions between objects and the background. Conversely, methods like BoxDiff and layout-control, despite offering more coherent image presentations, tend to suffer from issues such as missing objects or the appearance of unintended objects. In contrast, our proposed CSG method successfully combines precise layout placement with semantically coherent image quality, addressing these shortcomings (More qualitative comparisons can be found in supplementary materials).

\subsection{Quantitative analysis}

In a quantitative evaluation against current training-free techniques such as MultiDiffusion \cite{bar2023multidiffusion}, BoxDiff \cite{xie2023boxdiff}, and Layout-control \cite{chen2024training}, using Stable Diffusion version 1.4 \cite{Rombach_2022_CVPR}, our approach surpasses these methods in metrics of AP$_{50}$, AP, and CLIP score across datasets with 2, 3, and 2-4 objects, according to table \ref{tab:quant_results}. Compared to MultiDiffusion, which demonstrates strong object placement and recognition with high AP$_{50}$ and AP scores, it falls short in generating semantically meaningful content, as indicated by its lower CLIP score. Our method, however, not only improves upon these AP scores by 7.9, 6.6, and 5.8 respectively and AP scores by 1.9, 0.2, and 0.7, but also leads in CLIP score improvements by 0.0281 (7.3\%), 0.0217 (7.16\%), and 0.0206 (6.86\%). When compared to BoxDiff, which shows comparable performance in CLIP scores, our method demonstrates superior AP improvements by 12.5, 11.8, and 6.1 and AP$_{50}$ enhancements by 5.0, 3.2, and 5.6. These outcomes confirm the effectiveness of our method in achieving both accurate object generation and placement as well as semantic fidelity.

\begin{table}[H]
    \centering
    \small
    \setlength{\tabcolsep}{1mm}
    \renewcommand{\arraystretch}{1.0}
    \caption{Quantitative results on datasets with 2, 3 and 2-4 objects for comparison with concurrent training-free methods, including MultiDiffusion \cite{bar2023multidiffusion}, BoxDiff \cite{xie2023boxdiff} and Layout-control \cite{chen2024training}.}
    \label{tab:quant_results}
    \begin{tabular}{c|ccc|ccc|ccc}
        \toprule
        \multirow{2}{*}{Methods} & \multicolumn{3}{c|}{2 objects}& \multicolumn{3}{c|}{3 objects}&\multicolumn{3}{c} {2-4 objects}\\
         & AP$_{50}$ & AP & CLIP & AP$_{50}$ & AP & CLIP&AP$_{50}$&AP&CLIP\\
        \midrule
        Layout-control \cite{chen2024training}&24.4&7.84&0.3133&17.5&4.6&0.3181&23.1&7.01&0.3140\\
        BoxDiff \cite{xie2023boxdiff}&37.8&15.4&0.3124&30.3&11.3&0.3177&33.9&13.5&0.3132\\
        MultiDiffusion \cite{bar2023multidiffusion}&42.4&18.5&0.2983&35.5&14.3&0.3029&42.2&18.4&0.3003\\
        CSG (Ours)&\textbf{50.3}&\textbf{20.4}&\textbf{0.3201}&\textbf{42.1}&\textbf{14.5}&\textbf{0.3246}&\textbf{48.0}&\textbf{19.1}&\textbf{0.3209}\\
        \bottomrule
    \end{tabular}
\end{table}

\section{Conclusion}
This paper introduces a training-free method that surpasses previous approaches in object placement and generation, effectively addressing issues of semantic overlaps and self-attention misalignment that were overlooked by prior research. We showcase artworks created by our method which composite several objects and backgrounds. Compared to these earlier methods, our approach demonstrates superior performance in assimilating layout information and generating high-fidelity images. Through extensive experiments, we show that our method significantly mitigates adversarial effects encountered during the diffusion process. Although our work currently utilizes only bounding boxes as layout information, the proposed method is designed to be compatible with various forms of layout data. Furthermore, given its training-free nature, it can be seamlessly adapted to enhance models pre-trained with layout information, promising improved outcomes. For further discussions, please refer to the supplementary material provided.

\section*{Acknowledgements}
This work was supported in part by the Australian Research Council under Projects DP210101859 and FT230100549.

%
%
\bibliographystyle{splncs04}
\bibliography{main}

\begin{thebibliography}{10}
\providecommand{\url}[1]{\texttt{#1}}
\providecommand{\urlprefix}{URL }
\providecommand{\doi}[1]{https://doi.org/#1}

\bibitem{azizi2023synthetic}
Azizi, S., Kornblith, S., Saharia, C., Norouzi, M., Fleet, D.J.: Synthetic data from diffusion models improves imagenet classification. arXiv preprint arXiv:2304.08466  (2023)

\bibitem{balaji2022ediffi}
Balaji, Y., Nah, S., Huang, X., Vahdat, A., Song, J., Kreis, K., Aittala, M., Aila, T., Laine, S., Catanzaro, B., et~al.: ediffi: Text-to-image diffusion models with an ensemble of expert denoisers. arXiv preprint arXiv:2211.01324  (2022)

\bibitem{bar2023multidiffusion}
Bar-Tal, O., Yariv, L., Lipman, Y., Dekel, T.: Multidiffusion: Fusing diffusion paths for controlled image generation  (2023)

\bibitem{brock2018large}
Brock, A., Donahue, J., Simonyan, K.: Large scale gan training for high fidelity natural image synthesis. arXiv preprint arXiv:1809.11096  (2018)

\bibitem{cao2023masactrl}
Cao, M., Wang, X., Qi, Z., Shan, Y., Qie, X., Zheng, Y.: Masactrl: Tuning-free mutual self-attention control for consistent image synthesis and editing. arXiv preprint arXiv:2304.08465  (2023)

\bibitem{chai2023layoutdm}
Chai, S., Zhuang, L., Yan, F.: Layoutdm: Transformer-based diffusion model for layout generation. In: Proceedings of the IEEE/CVF Conference on Computer Vision and Pattern Recognition. pp. 18349--18358 (2023)

\bibitem{chefer2023attend}
Chefer, H., Alaluf, Y., Vinker, Y., Wolf, L., Cohen-Or, D.: Attend-and-excite: Attention-based semantic guidance for text-to-image diffusion models. ACM Transactions on Graphics (TOG)  \textbf{42}(4),  1--10 (2023)

\bibitem{chen2024training}
Chen, M., Laina, I., Vedaldi, A.: Training-free layout control with cross-attention guidance. In: Proceedings of the IEEE/CVF Winter Conference on Applications of Computer Vision. pp. 5343--5353 (2024)

\bibitem{croitoru2023diffusion}
Croitoru, F.A., Hondru, V., Ionescu, R.T., Shah, M.: Diffusion models in vision: A survey. IEEE Transactions on Pattern Analysis and Machine Intelligence  (2023)

\bibitem{dehouche2023s}
Dehouche, N., Dehouche, K.: What’s in a text-to-image prompt? the potential of stable diffusion in visual arts education. Heliyon  (2023)

\bibitem{dhariwal2021diffusion}
Dhariwal, P., Nichol, A.: Diffusion models beat gans on image synthesis. Advances in neural information processing systems  \textbf{34},  8780--8794 (2021)

\bibitem{ding2023temporal}
Ding, G., Sener, F., Yao, A.: Temporal action segmentation: An analysis of modern techniques. IEEE Transactions on Pattern Analysis and Machine Intelligence  (2023)

\bibitem{dosovitskiy2020image}
Dosovitskiy, A., Beyer, L., Kolesnikov, A., Weissenborn, D., Zhai, X., Unterthiner, T., Dehghani, M., Minderer, M., Heigold, G., Gelly, S., et~al.: An image is worth 16x16 words: Transformers for image recognition at scale. arXiv preprint arXiv:2010.11929  (2020)

\bibitem{gafni2022make}
Gafni, O., Polyak, A., Ashual, O., Sheynin, S., Parikh, D., Taigman, Y.: Make-a-scene: Scene-based text-to-image generation with human priors. In: European Conference on Computer Vision. pp. 89--106. Springer (2022)

\bibitem{goodfellow2014generative}
Goodfellow, I., Pouget-Abadie, J., Mirza, M., Xu, B., Warde-Farley, D., Ozair, S., Courville, A., Bengio, Y.: Generative adversarial nets. Advances in neural information processing systems  \textbf{27} (2014)

\bibitem{guo2020positive}
Guo, T., Xu, C., Huang, J., Wang, Y., Shi, B., Xu, C., Tao, D.: On positive-unlabeled classification in gan. In: Proceedings of the ieee/cvf conference on computer vision and pattern recognition. pp. 8385--8393 (2020)

\bibitem{hertz2022prompt}
Hertz, A., Mokady, R., Tenenbaum, J., Aberman, K., Pritch, Y., Cohen-Or, D.: Prompt-to-prompt image editing with cross attention control. arXiv preprint arXiv:2208.01626  (2022)

\bibitem{ho2020denoising}
Ho, J., Jain, A., Abbeel, P.: Denoising diffusion probabilistic models. Advances in neural information processing systems  \textbf{33},  6840--6851 (2020)

\bibitem{ho2022classifier}
Ho, J., Salimans, T.: Classifier-free diffusion guidance. arXiv preprint arXiv:2207.12598  (2022)

\bibitem{huang2024active}
Huang, T., Liu, J., You, S., Xu, C.: Active generation for image classification. arXiv preprint arXiv:2403.06517  (2024)

\bibitem{karras2019style}
Karras, T., Laine, S., Aila, T.: A style-based generator architecture for generative adversarial networks. In: Proceedings of the IEEE/CVF conference on computer vision and pattern recognition. pp. 4401--4410 (2019)

\bibitem{kim2023dense}
Kim, Y., Lee, J., Kim, J.H., Ha, J.W., Zhu, J.Y.: Dense text-to-image generation with attention modulation. In: Proceedings of the IEEE/CVF International Conference on Computer Vision. pp. 7701--7711 (2023)

\bibitem{li2023gligen}
Li, Y., Liu, H., Wu, Q., Mu, F., Yang, J., Gao, J., Li, C., Lee, Y.J.: Gligen: Open-set grounded text-to-image generation. In: Proceedings of the IEEE/CVF Conference on Computer Vision and Pattern Recognition. pp. 22511--22521 (2023)

\bibitem{li2023divide}
Li, Y., Keuper, M., Zhang, D., Khoreva, A.: Divide \& bind your attention for improved generative semantic nursing. arXiv preprint arXiv:2307.10864  (2023)

\bibitem{li2021image}
Li, Z., Wu, J., Koh, I., Tang, Y., Sun, L.: Image synthesis from layout with locality-aware mask adaption. In: Proceedings of the IEEE/CVF International Conference on Computer Vision. pp. 13819--13828 (2021)

\bibitem{lin2014microsoft}
Lin, T.Y., Maire, M., Belongie, S., Hays, J., Perona, P., Ramanan, D., Doll{\'a}r, P., Zitnick, C.L.: Microsoft coco: Common objects in context. In: Computer Vision--ECCV 2014: 13th European Conference, Zurich, Switzerland, September 6-12, 2014, Proceedings, Part V 13. pp. 740--755. Springer (2014)

\bibitem{liu2023diffusion}
Liu, D., Li, Q., Dinh, A.D., Jiang, T., Shah, M., Xu, C.: Diffusion action segmentation. In: Proceedings of the IEEE/CVF International Conference on Computer Vision. pp. 10139--10149 (2023)

\bibitem{liu2023application}
Liu, M., Hu, Y.: Application potential of stable diffusion in different stages of industrial design. In: International Conference on Human-Computer Interaction. pp. 590--609. Springer (2023)

\bibitem{parmar2023zero}
Parmar, G., Kumar~Singh, K., Zhang, R., Li, Y., Lu, J., Zhu, J.Y.: Zero-shot image-to-image translation. In: ACM SIGGRAPH 2023 Conference Proceedings. pp. 1--11 (2023)

\bibitem{radford2021learning}
Radford, A., Kim, J.W., Hallacy, C., Ramesh, A., Goh, G., Agarwal, S., Sastry, G., Askell, A., Mishkin, P., Clark, J., et~al.: Learning transferable visual models from natural language supervision. In: International conference on machine learning. pp. 8748--8763. PMLR (2021)

\bibitem{ramesh2022hierarchical}
Ramesh, A., Dhariwal, P., Nichol, A., Chu, C., Chen, M.: Hierarchical text-conditional image generation with clip latents. arXiv preprint arXiv:2204.06125  \textbf{1}(2), ~3 (2022)

\bibitem{ramesh2021zero}
Ramesh, A., Pavlov, M., Goh, G., Gray, S., Voss, C., Radford, A., Chen, M., Sutskever, I.: Zero-shot text-to-image generation. In: International Conference on Machine Learning. pp. 8821--8831. PMLR (2021)

\bibitem{Rombach_2022_CVPR}
Rombach, R., Blattmann, A., Lorenz, D., Esser, P., Ommer, B.: High-resolution image synthesis with latent diffusion models. In: Proceedings of the IEEE/CVF Conference on Computer Vision and Pattern Recognition (CVPR). pp. 10684--10695 (June 2022)

\bibitem{rombach2022high}
Rombach, R., Blattmann, A., Lorenz, D., Esser, P., Ommer, B.: High-resolution image synthesis with latent diffusion models. In: Proceedings of the IEEE/CVF conference on computer vision and pattern recognition. pp. 10684--10695 (2022)

\bibitem{saharia2022photorealistic}
Saharia, C., Chan, W., Saxena, S., Li, L., Whang, J., Denton, E.L., Ghasemipour, K., Gontijo~Lopes, R., Karagol~Ayan, B., Salimans, T., et~al.: Photorealistic text-to-image diffusion models with deep language understanding. Advances in Neural Information Processing Systems  \textbf{35},  36479--36494 (2022)

\bibitem{schuhmann2022laion}
Schuhmann, C., Beaumont, R., Vencu, R., Gordon, C., Wightman, R., Cherti, M., Coombes, T., Katta, A., Mullis, C., Wortsman, M., et~al.: Laion-5b: An open large-scale dataset for training next generation image-text models. Advances in Neural Information Processing Systems  \textbf{35},  25278--25294 (2022)

\bibitem{shi2020improving}
Shi, Z., Zhou, X., Qiu, X., Zhu, X.: Improving image captioning with better use of captions. arXiv preprint arXiv:2006.11807  (2020)

\bibitem{sohl2015deep}
Sohl-Dickstein, J., Weiss, E., Maheswaranathan, N., Ganguli, S.: Deep unsupervised learning using nonequilibrium thermodynamics. In: International conference on machine learning. pp. 2256--2265. PMLR (2015)

\bibitem{tumanyan2023plug}
Tumanyan, N., Geyer, M., Bagon, S., Dekel, T.: Plug-and-play diffusion features for text-driven image-to-image translation. In: Proceedings of the IEEE/CVF Conference on Computer Vision and Pattern Recognition. pp. 1921--1930 (2023)

\bibitem{vaswani2017attention}
Vaswani, A., Shazeer, N., Parmar, N., Uszkoreit, J., Jones, L., Gomez, A.N., Kaiser, L., Polosukhin, I.: Attention is all you need. Advances in neural information processing systems  \textbf{30} (2017)

\bibitem{wang2023yolov7}
Wang, C.Y., Bochkovskiy, A., Liao, H.Y.M.: {YOLOv7}: Trainable bag-of-freebies sets new state-of-the-art for real-time object detectors. In: Proceedings of the IEEE/CVF Conference on Computer Vision and Pattern Recognition (CVPR) (2023)

\bibitem{xie2023boxdiff}
Xie, J., Li, Y., Huang, Y., Liu, H., Zhang, W., Zheng, Y., Shou, M.Z.: Boxdiff: Text-to-image synthesis with training-free box-constrained diffusion. In: Proceedings of the IEEE/CVF International Conference on Computer Vision. pp. 7452--7461 (2023)

\bibitem{zhang2023adding}
Zhang, L., Rao, A., Agrawala, M.: Adding conditional control to text-to-image diffusion models. In: Proceedings of the IEEE/CVF International Conference on Computer Vision. pp. 3836--3847 (2023)

\bibitem{zheng2023layoutdiffusion}
Zheng, G., Zhou, X., Li, X., Qi, Z., Shan, Y., Li, X.: Layoutdiffusion: Controllable diffusion model for layout-to-image generation. In: Proceedings of the IEEE/CVF Conference on Computer Vision and Pattern Recognition. pp. 22490--22499 (2023)

\end{thebibliography}

\newpage

\appendix

\section{Experiment implementation}

The specifics of hyperparameter settings are discussed in previous sections. This section, however, concentrates on the rationale behind the experimental setup. Initially, we decided to omit any classes from the COCO 2014 dataset that consist of more than one word. This decision was made because many of these multi-word classes describe attributes rather than objects, such as "hot dog" and "potted plant." Our method is not designed to handle attribute association, hence we exclude these terms. The choice to utilize all classes for object detection with YOLOv7 stems from the observation that certain concepts within the class set bear similarities. For instance, attempting to generate an image of a "remote" might inadvertently produce an image of a "cell phone," a category we opted to exclude. Although this approach might compromise the method's performance in evaluations, it institutes a more rigorous assessment framework that more accurately reflects the method's capabilities.

\section{Analysis of inference time}

Inference time is evaluated against various baselines, and an ablation study demonstrates the impact of $T_D$ and $T_R$ on generation quality. The inference time is specifically measured for generating three objects. Although CSG lags behind other baselines in terms of speed, except for BoxDiff, this is because other methods either use a forward strategy or impose limited constraints during backward guidance.

\begin{table}[h]
    \centering
    \small
    \setlength{\tabcolsep}{3mm}
    \renewcommand{\arraystretch}{1.2}
    \caption{Inference time on datasets with 3 objects.}
    \begin{tabular}{c|ccccc}
        \toprule
        Methods & SD & Layout & BoxDiff & MultiDiff & CSG (Ours)\\
        \midrule
        Time (s) & 2.89 & 10.91 & 25.38 & 4.59 & 17.73\\
        \bottomrule
    \end{tabular}
\end{table}

However, as shown by out quantitative results, despite their faster inference speeds, MultiDiff and Layout-control exhibit limited performance. Our method outperforms these approaches by 18.6\% and 140.6\% on AP$_{50}$, respectively. While both CSG and BoxDiff involve calculating three losses, CSG achieves faster speeds while delivering superior performance.

\begin{table}[t]
    \centering
    \begin{minipage}[c]{0.49\linewidth}
        \small
        \setlength{\tabcolsep}{2mm}
        \renewcommand{\arraystretch}{1.2}
        \caption{Ablation study on $T_R$ with $T_D$ = 25.}
        \label{tb:ab_tr}
        \begin{tabular}{c|cccc}
        \toprule
            $T_R$ & AP$_{50}$ & AP & CLIP & Time (s)\\
            \midrule
            1 & 27.5 & 9.37 & 0.319 & 5.9 \\
            3 & 41.3 & 16.3 & 0.320 & 11.4\\
            5 & 50.3 & 20.4 & 0.320 & 16.5 \\
            7 & 51.1 & 21.6 & 0.315 & 23.4\\
            \bottomrule
        \end{tabular}
    \end{minipage}\hfill
    \centering
    \begin{minipage}[c]{0.49\linewidth}
        \small
        \setlength{\tabcolsep}{2mm}
        \renewcommand{\arraystretch}{1.2}
        \caption{Ablation study on $T_D$ with $T_R$ = 5.}
        \label{tb:ab_td}
        \begin{tabular}{c|cccc}
        \toprule
            $T_D$ & AP$_{50}$ & AP & CLIP & Time (s)\\
            \midrule
            15 & 44.8 & 18.5 & 0.318 & 11.2 \\
            20 & 47.4 & 20.3 & 0.318 & 14.4\\
            25 & 50.3 & 20.4 & 0.320 & 16.5 \\
            30 & 48.3 & 20.7 & 0.316 & 19.7\\
            \bottomrule
        \end{tabular}
    \end{minipage}
\end{table}

According to the results, increasing $T_D$ and $T_R$ leads to longer inference times. Balancing computational cost and performance is essential when selecting these parameters. Notable performance gains are observed up to $T_R$ = 5. Despite concerns that larger $T_R$ might cause the generation to converge to similar states, the high-dimensional latent space, model complexity, and limited refinement steps ensure diverse outputs. Cross-attentions guide the diffusion process during the first half (25 steps in our case). Exceeding this checkpoint results in longer inference times and decreased performance.

\section{Analysis of image quality}

In addition to the CLIP score, we also use FID and a user study to assess image quality. We compute the FID against images generated by SD using the same prompts but without layouts. The overall trend shows that incorporating more components results in a lower FID, indicating a divergence from the original distribution. However, this loss is compensated by a significant improvement in AP and AP$_{50}$. With all components activated, our method achieves a similar FID to the best baseline, indicating a comparable generation distribution. Relying on a single metric does not fully capture generation quality, as factors such as object location, object generation quality, and diversity compared to the base model must also be considered. Despite this, our method still outperforms other baselines according to a thorough quantitative analysis.

\begin{table}[H]
    \centering
        \small
        \setlength{\tabcolsep}{3mm}
        \renewcommand{\arraystretch}{1.2}
        \caption{Ablation study on proposed components measuring FID.}
        \label{tab:ab_guidance_fid}
        \begin{tabular}{ccccc|ccc}
            \toprule
            $\mathcal{L}_{intra}$ & $\mathcal{L}_{inter}$ & $\mathcal{L}_{self}$ & SS & AR & FID\\
            \midrule
            \checkmark &&& \checkmark & \checkmark & 51.52\\
            \checkmark & \checkmark &&&& 53.10\\
            \checkmark & \checkmark && \checkmark & \checkmark & 59.60\\
            \checkmark & \checkmark && \checkmark && 53.16\\
            \checkmark & \checkmark &&& \checkmark & 59.60\\
            \checkmark & \checkmark & \checkmark & \checkmark &  & 58.52\\
            \checkmark & \checkmark & \checkmark & \checkmark & \checkmark & 57.08\\
            \bottomrule
        \end{tabular}
\end{table}
\begin{table}[h]
    \centering
        \small
        \setlength{\tabcolsep}{3mm}
        \renewcommand{\arraystretch}{1.2}
        \caption{FID on datasets with 3 objects.}
        \label{tab:fid}
        \begin{tabular}{c|cccc}
            \toprule
            Methods & Layout & BoxDiff & MultiDiff  & CSG (Ours)\\
            \midrule
            FID & 56.32 & 56.00 & 61.27 & 57.08\\
            \bottomrule
        \end{tabular}
\end{table}

We conducted a user study with 20 sets of images, each containing 4 images (one for each method). All images shared the same layout, prompt, and seed. Forty users from a third-party labeling service were asked to choose the best-quality image from each set among images where all objects were correctly generated. Our method received the highest user preference.

\begin{table}[h]
    \centering
        \small
        \setlength{\tabcolsep}{3mm}
        \renewcommand{\arraystretch}{1.2}
        \caption{User study on randomly generated images.}
        \begin{tabular}{c|cccc}
            \toprule
             Methods & Layout & BoxDiff & MultiDiff & CSG (Ours)\\
            \midrule
            User Preference \% & 21.37 & 21.94 & 18.09 & 38.60\\
            \bottomrule
        \end{tabular}
\end{table}

\section{Performance with more objects}

We test our method on more challenging tasks involving the generation of five objects. The dataset used is a combined layout from COCO 2014. First, we collect all bounding boxes for each object. Then, we calculate the coexistence rate between each object. We randomly choose one object along with four others from the top ten objects with the highest coexistence rates. For each object, we select a bounding box that has less than 30\% overlap with the already selected bounding boxes.

\begin{table}[h]
    \centering
        \small
        \setlength{\tabcolsep}{3mm}
        \renewcommand{\arraystretch}{1.2}
        \caption{Quantitative results on datasets with 5 objects.}
        \begin{tabular}{c|cccc}
            \toprule
             Methods & Layout & BoxDiff & MultiDiff & CSG (Ours)\\
             \midrule
             AP$_{50}$ & 6.8 & 19.9 & 20.5 & 25.5\\
             AP & 1.81 & 6.55 & 7.58 & 7.82\\
             CLIP & 0.316 & 0.318 & 0.297 & 0.330\\
             \bottomrule
        \end{tabular}
\end{table}

Despite all methods experiencing a significant performance drop, our method still outperforms other baselines. It is important to note that training-free methods rely heavily on the model's generative abilities. As the number of objects increases, these methods may collapse, resulting in meaningless content.

\section{Compatibility with pre-trained layout-to-image model}

GLIGEN easily outperforms state-of-the-art training-free methods in layout-to-image synthesis. In practice, current training-free approaches typically pair with general conditioning models. The table shows that CSG, although not specifically designed for GLIGEN (which may lead to unnecessary computations and suboptimal performance), still significantly enhances performance. Moreover, these performance gains become more pronounced as generation tasks become more challenging.

\begin{table}[h]
    \centering
        \small
        \setlength{\tabcolsep}{2mm}
        \renewcommand{\arraystretch}{1.2}
        \caption{Quantitative results on datasets with 3 and 5 objects with GLIGEN.}
        \begin{tabular}{c|cccc|cccc}
            \toprule
            \multirow{2}{*}{Methods} & \multicolumn{4}{c|}{3 objects} & \multicolumn{4}{c}{5 objects}\\
             & AP$_{50}$ & AP & CLIP & Time & AP$_{50}$ & AP & CLIP & Time\\
            \midrule
            Gligen & 74.1 & 49.6 & 0.325  & 6.6 & 54.2 & 32.4 & 0.326 & 6.6\\
            Gligen+CSG & 76.8 & 51.6 & 0.326 & 19.6 & 60.6 & 34.3 & 0.336 & 21.63\\
            \bottomrule
        \end{tabular}
\end{table}

\section{Limitations}
Our method is effective at creating composite scenes, yet it encounters several notable issues. Primarily, while it focuses on local coherence by enhancing communication between objects and their immediate surroundings, it falls short in ensuring global coherence. This discrepancy is evident in the visual examples provided; the waterfall in image (a) appears isolated, contrasting with its more integrated appearance in images (b) and (c), where it seamlessly merges with the sky. Similarly, the castle's integration into the scene feels more natural in images (b) and (c) than in image (a). Another issue arises as the complexity of the semantics in the prompt increases. Attributes that are not explicitly anchored through layout information tend to be incorrectly associated with unintended objects. For example, across all three images, the blooms are incorrectly attached to the unicorn, resulting in an unintended color scheme for the creature.

\begin{figure*}[ht]
    \centering
    \includegraphics[width=\linewidth]{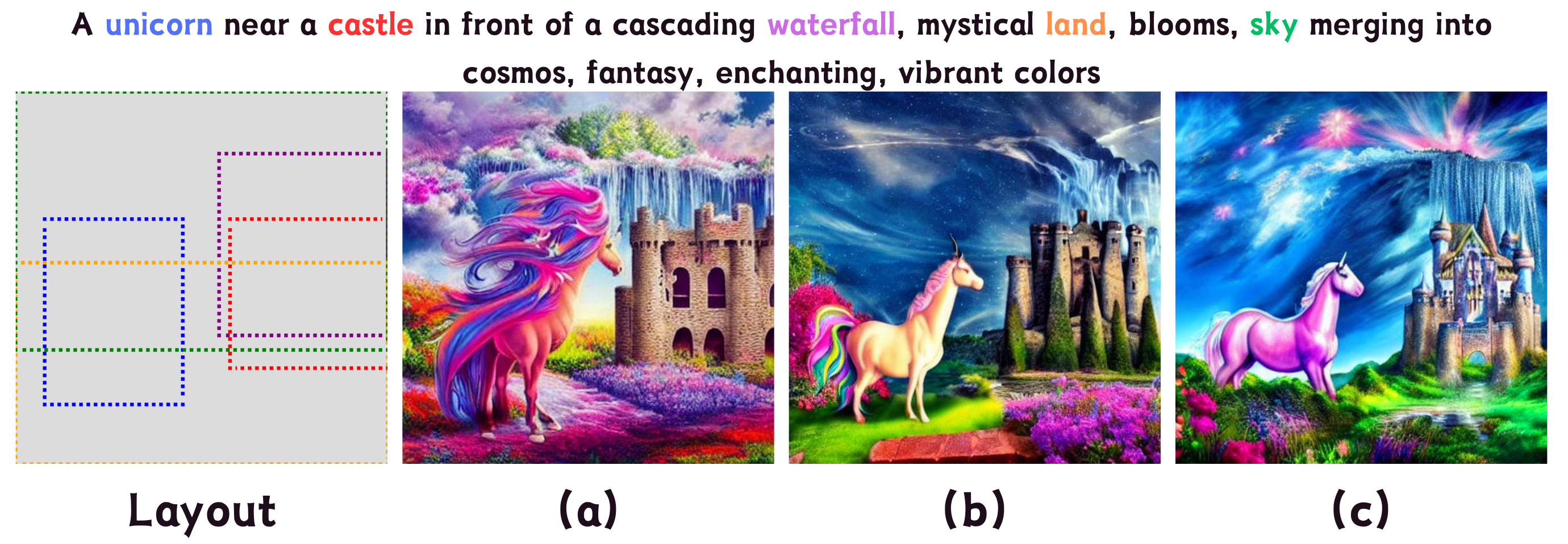}
    \caption{The limitation lies in the globally incoherent and incorrectly attributed binding of synthesized images.}
    \label{fig:limit}
\end{figure*}

\section{More visual evaluation}

We offer additional visual comparisons between our method and other training-free approaches using the COCO 2014 dataset with three objects. We also provide visualizations of our customized scenes.

\begin{figure*}[h]
    \centering
    \includegraphics[width=\linewidth]{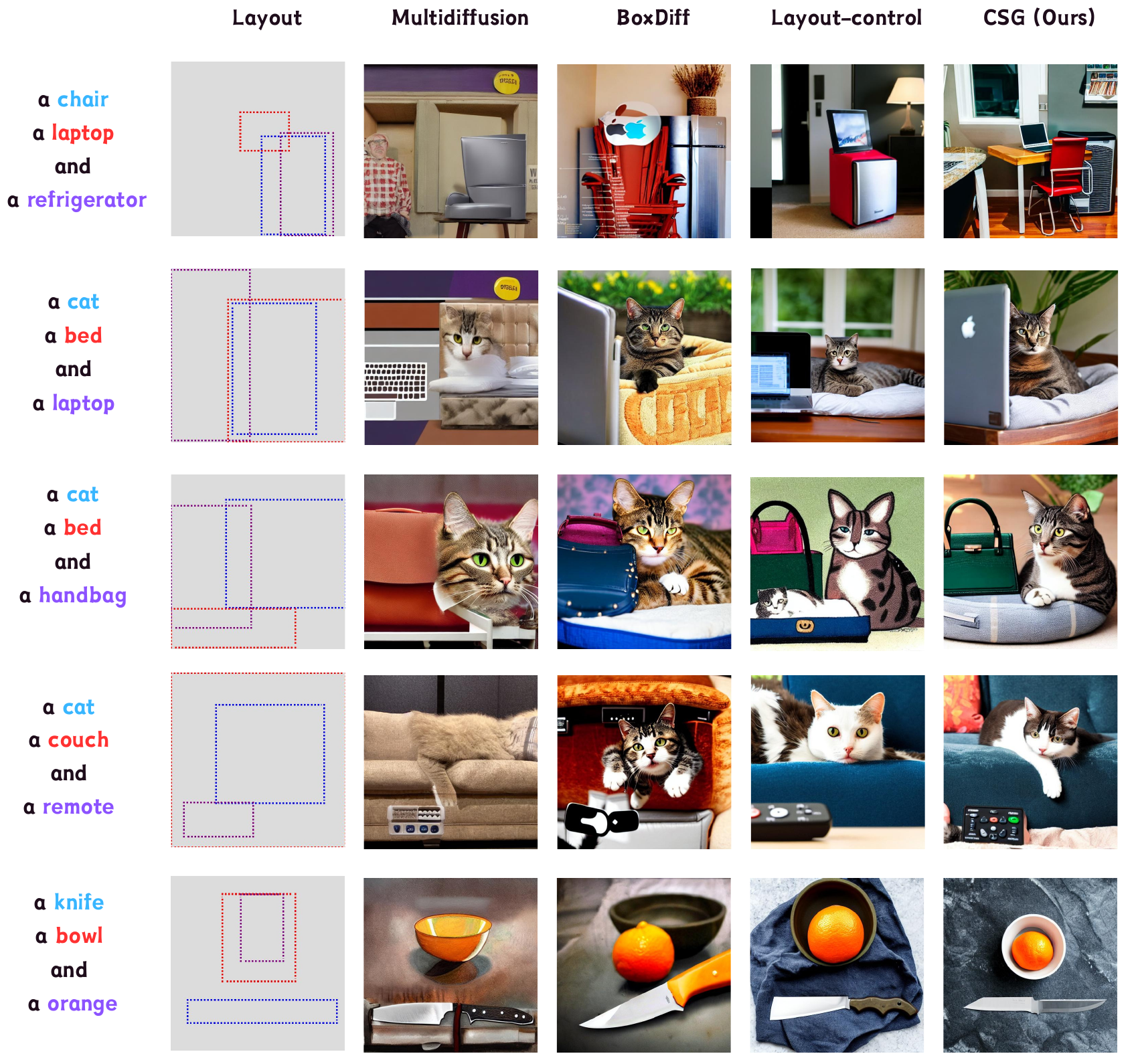}
    \caption{Extra visual comparison on COCO 2014 with 3 objects.}
    \label{fig:comp_ext}
\end{figure*}

\begin{figure*}[ht]
    \centering
    \includegraphics[width=\linewidth]{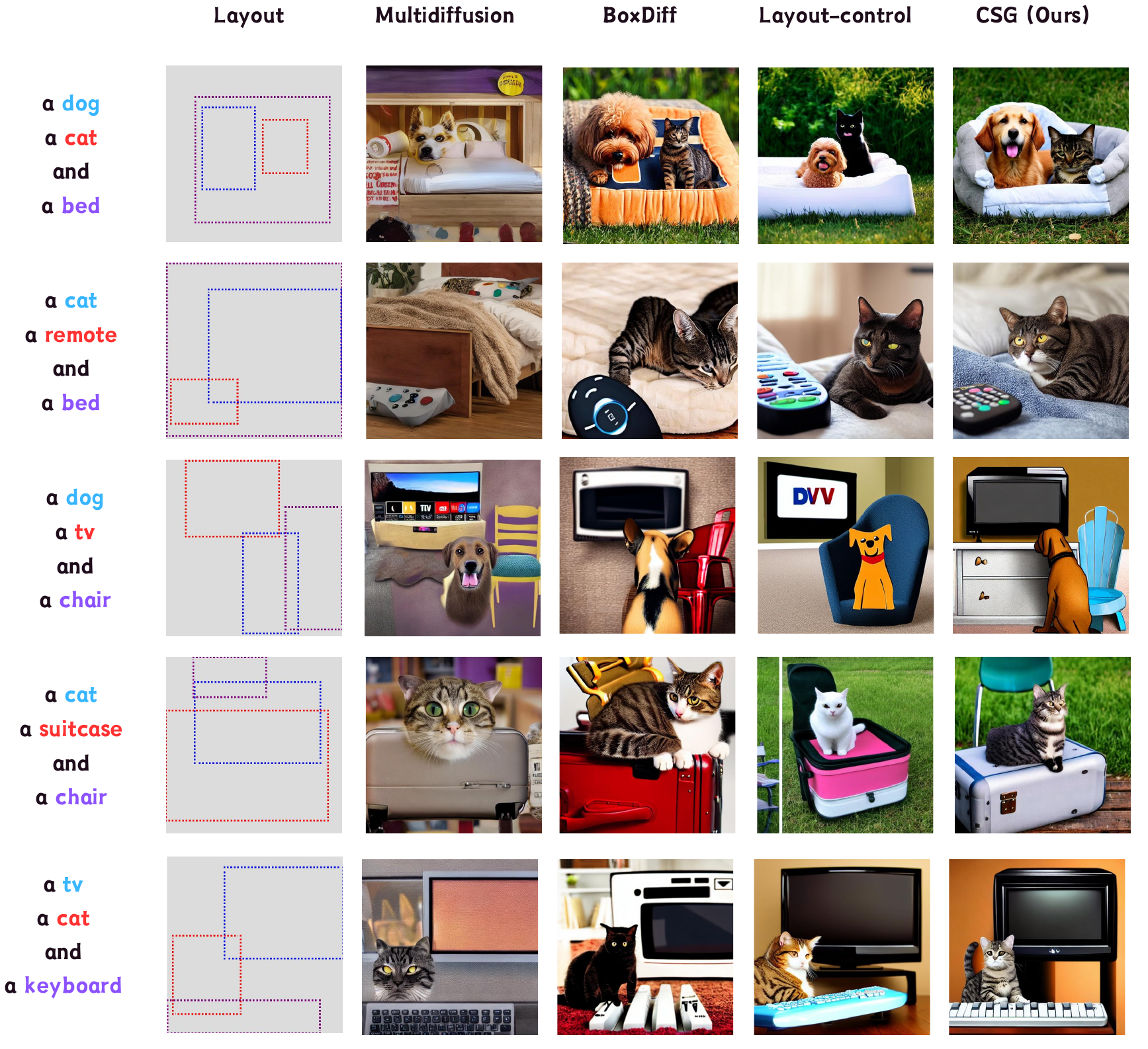}
    \caption{Extra visual comparison on COCO 2014 with 3 objects cont.}
    \label{fig:comp_ext_cont}
\end{figure*}

\begin{figure*}[ht]
    \centering
    \includegraphics[width=\linewidth]{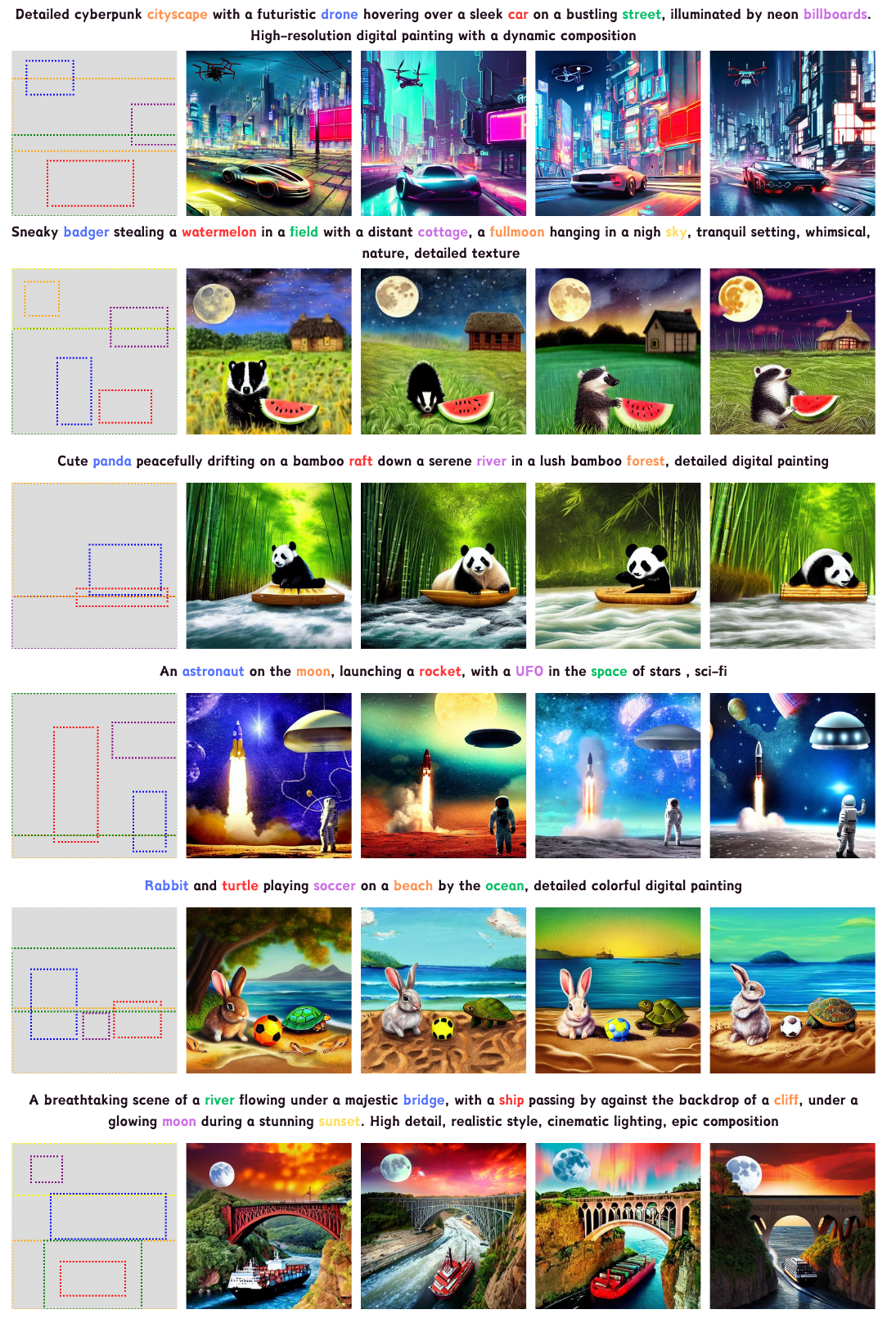}
    \caption{Extra visualization of customized scenes.}
    \label{fig:art}
\end{figure*}

\begin{figure*}[ht]
    \centering
    \includegraphics[width=0.98\linewidth]{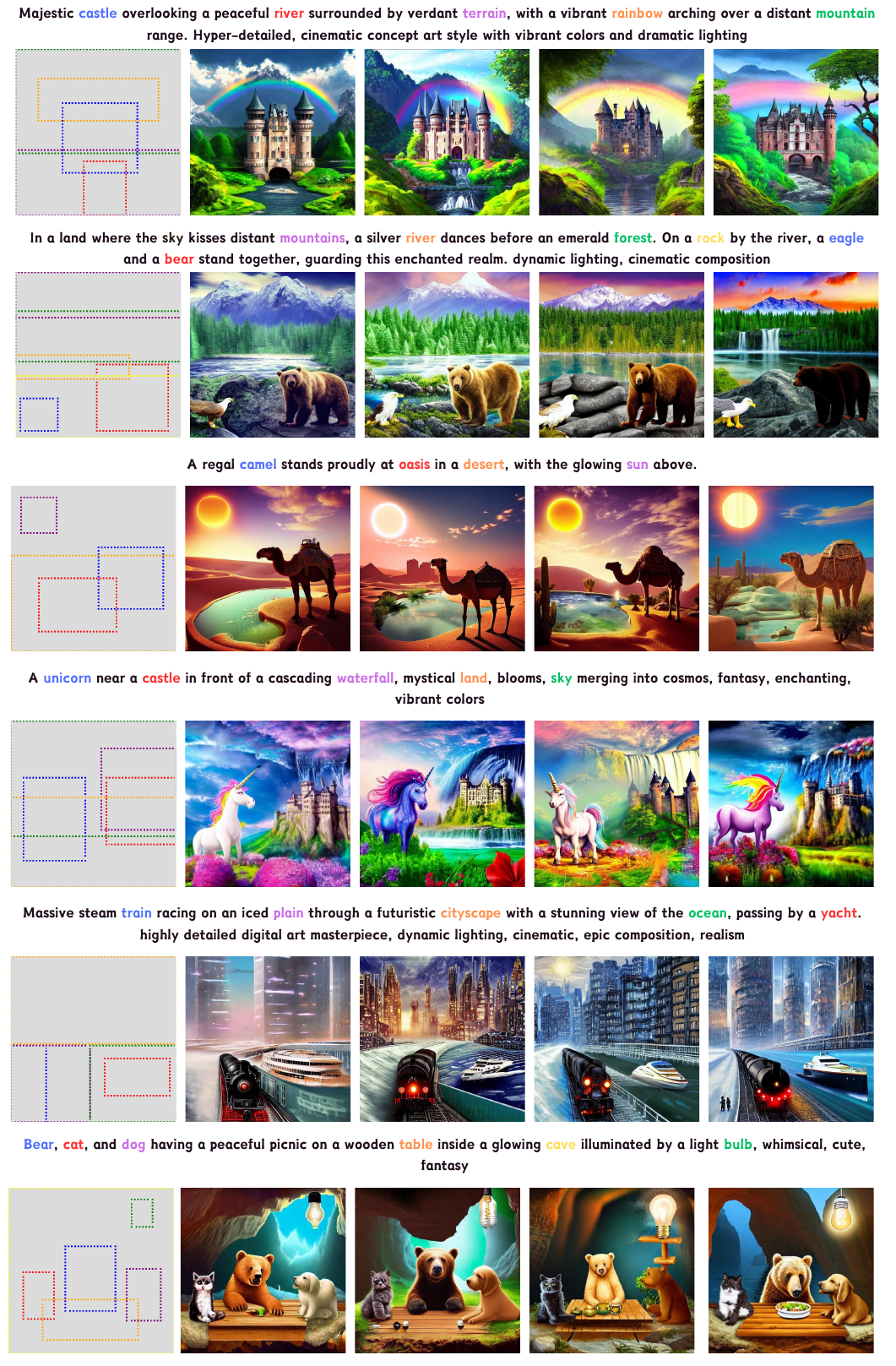}
    \caption{Extra visualization of customized scenes cont.}
    \label{fig:art_cont}
\end{figure*}

\end{document}